\documentclass{article}

\usepackage[final]{hrl_nips_2017}

\usepackage[utf8]{inputenc} % allow utf-8 input
\usepackage[T1]{fontenc}    % use 8-bit T1 fonts
\usepackage{hyperref}       % hyperlinks
\usepackage{url}            % simple URL typesetting
\usepackage{booktabs}       % professional-quality tables
\usepackage{amsfonts}       % blackboard math symbols
\usepackage{nicefrac}       % compact symbols for 1/2, etc.
\usepackage{microtype}      % microtypography

\usepackage{amsthm}
\usepackage{amssymb,amsmath,bm,booktabs}
\usepackage{amsfonts}
\usepackage[linesnumbered,ruled,vlined]{algorithm2e}
\usepackage{comment}
\usepackage{subfigure}
\usepackage[colorinlistoftodos]{todonotes}
\usepackage[authormarkuptext=name,addedmarkup=bf,authormarkupposition=left]{changes}
\definechangesauthor[name={M.~L.}, color={red}]{ml}
\definechangesauthor[name={S.~Y.}, color={blue}]{syi}   %cyan
\definechangesauthor[name={XT.~L.}, color={brown}]{xtLi}   %cyan
\definechangesauthor[name={P.~Chen}, color={cyan}]{cp}

\newcommand*\samethanks[1][\value{footnote}]{\footnotemark[#1]}
\title{The Eigenoption-Critic Framework}

\author{
\hspace{-0.2in}
  Miao Liu %\thanks{Address: T. J. Watson Research Center, 1101 Kitchawan Rd, Yorktown Heights, NY 10598}
   \\
\hspace{-0.2in}  IBM AI \\
\hspace{-0.2in}  \texttt{miao.liu1@ibm.com} \\
  \And
\hspace{-0.15in}  Marlos C. Machado \\
\hspace{-0.15in}  University of Alberta\\
\hspace{-0.15in}  \texttt{machado@ualberta.ca} \\
 \And
\hspace{-0.15in}  Gerald Tesauro\samethanks \\
\hspace{-0.15in}  IBM AI\\
\hspace{-0.15in}  \texttt{gtesauro@us.ibm.com} \\
  \And
\hspace{-0.15in}  Murray Campbell\samethanks \\
\hspace{-0.15in}  IBM AI\\
\hspace{-0.15in}  \texttt{mcam@us.ibm.com} \\
}

\begin{document}

\maketitle

\begin{abstract}
%\vskip-0.1in
  %This paper presents a novel algorithm termed eigen-option critic(EOC) for hierarchical reinforcement learning (HRL). EOC is an option-critic (OC) method augmented by eigen-purposes, which are intrinsic rewards defined by eigenvectors of the graph Laplacian of a state-space.
  %Option-critic (OC)~\cite{bacon2017option} is a general hierarchal reinforcement learning (HRL) architecture that allows learning the intra-option policies and termination functions, simultaneously with the policy over them. However, the existing OC learning method only relies on extrinsic rewards, hence there is no guarantee that the learned options are diverse and transferable. More recently, a graph Laplacian based approach has been proposed to discover eigenoptions (EO)~\cite{Machado2017ALF} which has shown to be diverse and to allow efficient exploration. However, the existing EO method requires two separate steps (eigenoption discovery and reward maximization) to learn a control policy, which can incur a significant amount of storage and computation, and is only designed for problems with discrete state-spaces. Here, we proposed a new HRL architecture termed eigenoption-critic (EOC), which augments OC with eigenpurposes, and allows learning of diversified options and policies over them simultaneously. Moreover, we generalize EOC to solving problems with continuous state-spaces using the Nystr\"{o}m approximation. Preliminary experimental results in both discrete and continuous domains under a nonstationary setting, demonstrate EOC outperforms OC in general. 
Eigenoptions (EOs) have been recently introduced as a promising idea for generating a diverse set of options through the graph Laplacian, having been shown to allow efficient exploration~\cite{Machado17a}. Despite its first initial promising results, a couple of issues in current algorithms limit its application, namely: 1)~EO methods require two separate steps (eigenoption discovery and reward maximization) to learn a control policy, which can incur a significant amount of storage and computation; 2)~EOs are only defined for problems with discrete state-spaces and; 3)~it is not easy to take the environment’s reward function into consideration when discovering EOs. In this paper, we introduce an algorithm termed eigenoption-critic (EOC) that addresses these issues. It is based on the Option-critic (OC) architecture~\cite{bacon2017option}, a general hierarchical reinforcement learning algorithm that allows learning the intra-option policies simultaneously with the policy over options. We also propose a generalization of EOC to problems with continuous state-spaces through the Nystr\"{o}m approximation. EOC can also be seen as extending OC to nonstationary settings, where the discovered options are not tailored for a single task.
\end{abstract}

\section{Introduction}
%\vskip-0.07in
Reinforcement Learning (RL) has been a driving force of many recent AI breakthroughs, such as systems capable of matching or surpassing human performance in \emph{Jeopardy!}~\cite{DBLP:journals/jair/TesauroGLFP13}, Go~\cite{SilverHuangEtAl16nature,silver2017mastering}, Atari games~\cite{mnih2015humanlevel}, autonomous driving~\cite{CADRL-ICRA17,SA-CADRL-IROS17} and conversing with humans~\cite{milabot}. In these domains, the RL policy makes step-by-step decisions, operating at the finest-grained %possible 
time scale. %We call these decisions primitive actions. For primitive-action based RL methods, 
Such an approach is not feasible in several other domains where the delayed credit assignment is a major challenge. A principled approach to combat this challenge is to use \emph{temporally extended} actions, allowing agents to operate at a higher level of abstraction by having actions being executed for different amounts of time. Temporally extended actions lead to hierarchical control architectures. They provide a divide-and-conquer approach to tackle complex tasks by learning to compose the right set of subroutines to succeed in the task. It also potentially allows much faster adaptation to changes in the environment, which may only involve small changes in the task hierarchy (\emph{e.g.}, changing the goal state in maze navigation task). %As such, tasks organized in hierarchical form may provide an important underpinning of success in lifelong learning scenarios, which require continual adaptation to ever-changing environments. 
In this work we are interested in developing efficient methods for automatically learning task hierarchies in RL. We focus on the options framework~\cite{Sutton99}, a general framework for modeling temporally extended actions.% Hierarchical RL (HRL)~\cite{BOTVINICK2012956}, which allows decision making to happen at different temporal scales has been shown to be a promising way for accelerating RL in lifelong learning. 
Autonomous option discovery has been subject of extensive research since the late 90's, with a large number of HRL algorithms being based on the options framework~\cite{Simsek04, Daniel16, Florensa17, Konidaris09, Machado17c, Mankowitz16, McGovern01}. In this paper we explore the recently introduced concept of eigenoptions (EOs)~\cite{Machado17a}, a  set of diverse options, obtained from a learned latent representation, that allows efficient exploration. Such an idea is promising and quite different from others in the literature. However, existing algorithms for eigenoption discovery have the following limitations: 1) They have two clear distinct phases that can incur significant storage and computational costs, namely (i)~learning the intra-option policies and (ii)~learning the policy over options to complete the task; 2) They are only defined for problems with discrete state-spaces and 3) they do not support the use of the environment's reward function in the eigenoption discovery process. In this paper we introduce an algorithm for eigenoption discovery that addresses these issues. We do so by porting the idea of eigenoptions to the the option-critic (OC) architecture~\cite{bacon2017option}, which is capable of simultaneously learning intra-option policies and the policy over options; and by using the Nystr\"{o}m approximation to generalize eigenoptions to problems with continuous state-spaces. We term our algorithm Eigenoption-critic (EOC). From the OC's perspective, our algorithm can be seen as extending the OC architecture to be able to better deal with nonstationary environments and to be able to generate more diverse options that are general enough to be transferable across tasks.
\section{Preliminaries and Notations}
%\vskip-0.07in
%\subsection{Reinforcement Learning and Options}
%Reinforcement learning (RL) is a class of machine learning methods for solving sequential decision making problems with unknown state-transition dynamics.
We consider sequential decision-making problems formulated as Markov decision processes (MDPs). An MDP is specified by a quintuple $\mathcal{M}=\langle \mathcal{S},\mathcal{A},T,R,\gamma \rangle$, where $\mathcal{S}$ is the state space, $\mathcal{A}$ is the action space, $T$ is the state-transition model, $R$ is the reward function, and $\gamma\in [0,1)$ is a discount factor. The goal of an RL agent is to find a control policy $\pi:\mathcal{S}\times\mathcal{A}\rightarrow[0,1]$ that maximizes the expected discounted return $G_t = \sum_{k=0}^\infty\gamma^t r_{k+t+1}$. The policy can be derived from a value function such as the state-value function $V^\pi(s)$ or the state-action value function $Q^\pi(s,a)$. Specifically, we have $V^\pi(s)=\mathbb{E}_{\pi,T}[G_t|s]=\sum_{a\in\mathcal{A}}\pi(a|s)Q^\pi(s,a)$. In RL settings we use samples collected through agent-environment interactions to estimate the value functions.
%Specifically, $Q^\pi(s,a)$ is defined as $Q^\pi(s,a) = \mathbb{E}\big[\sum_{t=1}^T\gamma^t r_t|s_0=s, a_0=a\big]\forall a\in\mathcal{A}, \forall s\in\mathcal{S}$ with $V^\pi(s')=\sum_{a'\in\mathcal{A}}\pi(a|s)Q^\pi(s',a')$. Note that the $Q$-functions have to be estimated based on samples collected through agent-environment interactions.

%\begin{equation}
%Q^\pi(s,a) = \mathbb{E}\big[\sum_{t=1}^T\gamma^t r_t|s_0=s, a_0=a\big]=\sum_{s'\in\mathcal{S}}P(s'|s,a)\bigg[R(s,a,s')+\gamma V^\pi(s')\bigg]ds',\;\; \forall a\in\mathcal{A}, \forall s\in\mathcal{S}, 
%\end{equation}
%with $V^\pi(s')=\int_{a'\in\mathcal{A}}\pi(a|s)Q^\pi(s',a')$. 

\textbf{Options} in HRL, are defined by a set of trituples $O=\{I_o, \beta_o, \pi_o\}_{o=1}^K$, where $K$ is the total number of options, $I_o \in \mathcal{S}$ is the initiation set of option $o$, $\beta_o: \mathcal{S} \rightarrow [0,1]$ is the stochastic termination condition of option $o$, and $\pi_o: \mathcal{S}\times \mathcal{A} \rightarrow [0,1]$ is the stochastic intra-option policy of option $o$.

\textbf{The Option-Critic architecture} is a policy gradient method that provides a general framework for option discovery in HRL~\cite{bacon2017option}.
%Unlike most existing work separating option discovery and learning, which incurs a significant amount of storage and computation.
The OC architecture facilitates joint option learning and automatic option discovery on the fly. Specifically, it is based on the following state-option value and state-action-option value:
%\vskip-0.35in
\begin{eqnarray}\label{eq:Qo}
&Q_O(s,o) = \sum_a \pi_o(a|s)Q_U(s,a, o), 
\\\label{eq:Qu}
&Q_U(s,a,o) = r(s,a) + \gamma\int_{s'}Pr(s'|s,a)U(s',o) 
\end{eqnarray}
where $U(s',o) = \big(1-\beta_o(s')\big)Q_O(s',o) + \beta_o(s')V(s')$. One can use the policy gradient theorem~\cite{Sutton:1999:PGM:3009657.3009806} to obtain the gradient of the expected discounted return with respect to the intra-option policy parameter $\theta$ and to the option termination function parameter $\eta$, respectively:
\begin{eqnarray}
\label{eq:intra-option-policy-gradient}
&\frac{\partial Q_O(s,o)}{\partial \theta}=\int_{s}\sum_{o}\mu_O(s,o|s_0, o_0)\sum_{a}\frac{\partial \pi_{o,\theta}}{\partial \theta}Q_U(s,o,a),
\\
\label{eq:option-termination-policy-gradient}
&\frac{\partial Q_O(s,o)}{\partial \eta}=-\int_{s'}\sum_{o}\mu_O(s,o|s_0, o_0)\frac{\partial\beta_{o}(s')}{\partial\eta}\big(Q_U(s,a,o) - V_\Omega(s')\big), 
\end{eqnarray}
where $\mu_O(s,o|s_1, o_0) = \sum_{t=0}^\infty \gamma^tP(s_{t+1}=s, o_t=o|s_1,o_0)$.  The OC architecture simultaneously learns intra-option policies, termination conditions, and the policy over options through two steps: 1)~\emph{Critic step}: evaluating $Q_O$~(Eq.~\ref{eq:Qo}) and $Q_U$~(Eq.~\ref{eq:Qu}) with TD errors; and 2)~\emph{Actor step}: estimating Eq.~\ref{eq:intra-option-policy-gradient} and Eq.~\ref{eq:option-termination-policy-gradient} to update the parameters of the termination functions and of the intra-option policies. Notice that the OC algorithm is a policy gradient method focused on maximizing the return. It does not try to capture the topology of the state-space and it does not incentivize the agent to keep exploring the environment. Thus, the OC can be inefficient for solving nonstationary tasks as the discovered options tend to be tailored for a particular task instead of being general for better transferability.

\section{The Eigenoption-Critic Framework}
%\vskip-0.07in
%\subsection{Eigen-Options}
Eigenoptions~\cite{Machado17a} are a set of options obtained from the eigenvectors of the graph Laplacian generated by the undirected graph formed by the state transitions induced by the MDP. Specifically, assume an adjacency matrix $W$ can be estimated through a Gaussian kernel with each element specified as $W_{i,j}=w(s_i,s_j) = \alpha\exp\{-|s_i - s_j|/\sigma\}$, upon which we can build a combinatorial graph Laplacian $L = D - W$, where $D$ is a diagonal matrix with the $i$-th diagonal element defined as $D_{i,i}=\sum_{j}W_{i,j}$. By performing eigenvalue decomposition of $L$ one obtains eigenvectors, also known in the RL community as proto-value functions (PVFs)~\cite{Mahadevan07}. The PVFs capture large-scale temporal properties of a diffusion process and have been shown to be useful for representation learning~\cite{mahadevan2007proto} and have been used as intrinsic reward signals for guiding exploration~\cite{Machado17a,Machado17c}. Specifically, given $e_o$, the $o$-th eigenvector of $L$, one can specify an intrinsic reward function as:
\begin{eqnarray}\label{eq:intrinsic_reward}
r_{in}(s,s',o) = e_o^T\big(\phi(s')-\phi(s)\big), 
\end{eqnarray}
where $\phi(\cdot)$ denotes the feature representation of a given state. This reward can be seen as an intrinsic motivation for the agent to explore the environment. Essentially, $L$ defines a diffusion model, which captures the information flow on a graph or a manifold (the topology of the underlying state space). %\ml{Need to make this more concrete.}

\begin{center}
%\vskip-0.11in
\scalebox{0.92}{
\begin{minipage}{1.0\linewidth}
\begin{algorithm}[H]\label{alg:EOC}
	\textbf{Input:} Reward mixing weight $\alpha$, learning rates $\alpha_\theta$, $\alpha_\theta$, $\alpha_o$, $\alpha_u$, discount factor $\gamma$\\
	\textbf{Output:} A set of parameterized options $O=\{I_o, \pi_o^\pi(a|s), \beta_o(s)\}$ and a policy over options $\pi(o|s)$ \\
	\For{i={$1,\ldots, N_{episodes}$}}{ 
		Choose option $o$ according to $\epsilon$-soft($\pi(o|s')$)\\
		\While{not reached goal}{ 
			Choose an action $a\sim \pi^b_{o, \theta}(a|s)$, and observe $s'$, $r_{ex}$ \\
	Find the k nearest neighbors of $s$, $\{s_1,...,s_k\}$\\
	\For{i={$1,\ldots, k$}}{ 
		compute $w(s, s_i)$}
	%Compute $d(s) = \sum w(s,s_i)$, 
	Compute eigenfunction value at $s$, intrinsic reward using and mixed reward using equations \eqref{eq:nystrom}, \eqref{eq:intrinsic_reward}, and \eqref{eq:combined_reward}, respectively\\
			%\HiLi $\xi \leftarrow \pi_{o, \theta}(a|s)/\pi_{o}^b(a|s)$\\
			\textbf{1. Update the critic (Option evaluation)}:\\
			%\HiLi $\bm e \leftarrow \xi(\bm \phi(s)+\gamma\lambda\bm e)$\\ 
			$\delta_u \leftarrow r - Q_U(s,o,a)$, $\delta_o \leftarrow r_{ex} - Q_O(s,o)$\\
			\If{$s'$ is not terminal}{
				$g \leftarrow \gamma(1-\beta_{o,\eta}(s'))Q_o(s',o) + \gamma\beta_{o,\eta}(s')\max_{o'}Q_\omega(s',o)$, 
				$\delta_u \leftarrow \delta_u + g$, $\delta_o \leftarrow \delta_o + g$
			} 
			$Q_U(s,o,a) \leftarrow Q_U(s,o,a) + \alpha_u\delta_u$, 
			$Q_O(o,s) \leftarrow Q_O(o,s) + \alpha_o\delta_o$\\
			%\HiLi $Q_U(s,o,a) \leftarrow Q_U(s,o,a) + \alpha_u\delta_u\bm e(s)$\\
			%\HiLi $Q_O(o,s) \leftarrow Q_O(o,s) + \alpha_o\delta_o\bm e(s)$\\
			\textbf{2. Update the actor (Options improvement)}:\\
			$\theta \leftarrow \theta + \alpha_\theta\frac{\partial\ln\pi_{o,\theta}(a|s)}{\partial\theta}Q_U(s,o,a)$, 
			 $\eta \leftarrow \eta -\alpha_\eta\frac{\partial\beta_{o,\eta}(s')}{\partial\eta}(Q_O(s',o)-V_O(s'))$\\
			\If{$\beta_{o,\eta}$ terminates in $s'$}{
				choose new option $o$ according to $\epsilon$-soft($\pi(o|s')$)
			}
			$s\leftarrow s'$
		}		
	}
	%\Return $O$
	\caption{Eigenoption-Critic with tabular intra-option Q learning}
\end{algorithm} 
\end{minipage}
}
\end{center}

\subsection{Eigenoption-Critic Framework}
%\vskip-0.07in
Previous work on eigenoptions~\cite{Machado17a} maximizes rewards in a two phase procedure: off-line eigenoption discovery and online reward maximization. % (primive-action based off-policy policy learning using learned eigenoption for exploration). 
In this work we propose to learn eigenoptions online by integrating the intrinsic reward into the OC architecture. By doing so the agent can learn eigenoptions while it does not observe extrinsic rewards, which can be interpreted as an auxiliary task~\cite{Jaderberg17}. This can also be seen as promoting exploration before the agent stumbles upon the goal state. We define the following mixed reward function as a convex combination of the extrinsic and intrinsic rewards:
%\vskip-0.2in
\begin{equation}\label{eq:combined_reward}
r(s,a,o) = \alpha r_{in}(s,s',o) + (1-\alpha) r_{ex}(s,a). 
\end{equation}
The motivation is that we learn options that encode information about diffusion, but do not ignore the extrinsic rewards. Note that we can recover the original options in OC and EOs when $\alpha$ is set to $0$ and $1$ respectively. Also, note that the mixed reward (Eq.~\ref{eq:combined_reward}) is only used for learning options, while the policy over options is still learned by extrinsic rewards, since it is the only useful feedback for evaluating task performance. Algorithm~1 summarizes the EOC algorithm in the tabular case.
%\ml{discuss more about the implications. This should be related to reward shaping techniques, check more details.} 

\subsection{Generalization of the Eigenoption-Critic Framework to Continuous Domains}
%\vskip-0.07in
In this section, we discuss the extension of the Eigenoption-critic framework to problems with continuous state-spaces. Although the OC architecture naturally handles continuous domains, the EO framework was not defined for that. We fill this gap by showing how one can generalize eigenvectors to eigenfunctions and how one can interpolate the value of eigenvectors computed on sampled states to a novel state. This generalization is achieved via the Nystr\"{o}m approximation~\cite{mahadevan2007proto},
%\vskip-0.2in
\begin{eqnarray}\label{eq:nystrom}
e_o(s) = \frac{1}{1-\lambda_o}\sum_{i : |s-s_i|<\epsilon}\frac{w(s,s_i)}{\sqrt{d(s)d(s_i)}} e_o(s_i)
\end{eqnarray}
where $\lambda_o$ is the $o$-th eigenvalue, $e_o(s_i)$ is the $o$-th eigenvector value at an anchor point $s_i$, $d(s) = \sum_{i:|s-s_i|<\epsilon} w(s,s_i)$ is the degree of $s$, a new vertex on the graph, and $d(s_i)$ is the degree of an anchor point $s_i$. Note that $d(s)$ is determined when a new state $s$ is encountered in the online learning phase, whereas $d(s_i)$ is precomputed when the graph is constructed. 
%\ml{discuss a little more about anchor point selection}
\begin{figure}[t]
	\begin{center}
		%\vspace{-12pt}
				%	\fbox{
	$\begin{array}{c}
	\hspace{-0.0cm}\includegraphics[scale=0.325]{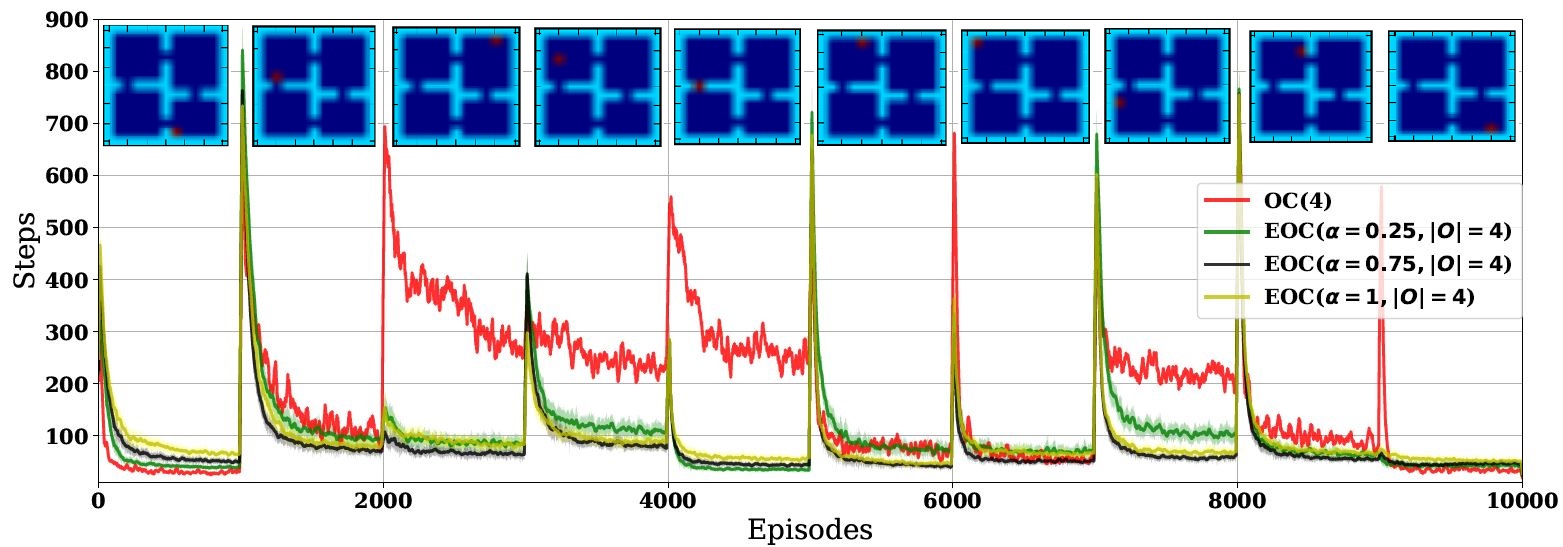}
	\end{array}$
	%\vskip -0.1in		%       			}
	\caption{\footnotesize{The performance of Algorithm~\ref{alg:EOC} using different reward mixing coefficients $\alpha$ on a benchmark: 4-rooms under a nonstationary setting, where the goal location changes periodically.}}
	\label{fig:performance}
	\end{center}
\end{figure}
\begin{figure}[t]
	\begin{center}
		%\vspace{-12pt}
				%	\fbox{
	$\begin{array}{c}
	 \includegraphics[scale=0.345]{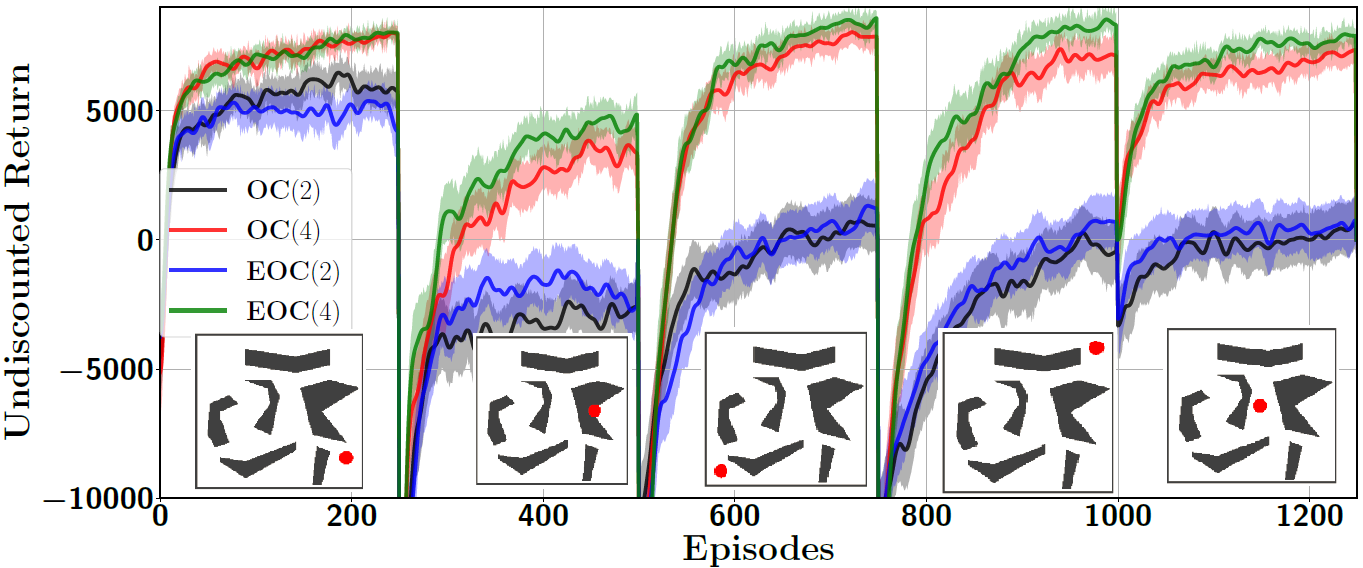}
	\end{array}$
	%\vskip -0.1in		%       			}
	\caption{\footnotesize{The performance comparison between OC and EOC ($\alpha=0.5$) on a benchmark: pinball under a nonstationary setting, where the goal location changes periodically.}}
	\label{fig:pinball}
	\end{center}
	\vskip -0.25in	
\end{figure}
\section{Experiments}
We validate the proposed method on two benchmarks under a nonstationary setting, where goal locations change periodically, which the agent is unaware of. We use the same parameters reported by Bacon et al.~\cite{bacon2017option} in our experiments. Due to space limit, readers are referred to their paper for details.   

\textbf{Four-rooms domain} The first experiment is performed on a navigation task in the 4-rooms domain~\cite{Sutton99}, which has discrete state-space. The initial state is randomly generated for each episode. After every $1,000$ episodes the goal location changes to a new one following a uniform distribution over valid positions. The performance is measured by the number of steps the agent takes to reach the goal, and is shown on Figure~\ref{fig:performance}. It can be seen that, for the first goal setting, the OC architecture outperforms other methods when enough samples are used. This is due to the fact that OC's learning mechanism is solely based on extrinsic rewards, which is only obtained when the agent hits the goal location. Whereas for EOC the intrinsic reward, which allows more efficient exploration, is also used for updating intra-option policies. When the goal location changes, because of more efficient exploration, EOC outperforms OC for most of the subsequent tasks, especially when the goal location changes between two rooms. EOC with $\alpha=0.75$ is the setting that achieves the best performance, suggesting that the combination of both approaches is indeed promising.  

\textbf{Pinball domain} The second experiment is performed on the pinball domain~\cite{Konidaris09}, which has a continuous state space. The agent starts in the upper left corner in every episode. The goal location changes after 250 episodes (red circle in Figure~\ref{fig:pinball}). In this domain, we used intra-option Q-learning with linear function approximation over Fourier basis. We used the data collected from the first 10 episodes to build a Kd-tree and we search k=15 nearest neighbors to perform the Nystr\"{o}m approximation of the eigenfunctions. We plot the undiscounted return on Figure~\ref{fig:performance}. Similar to the first experiment, except for the first task, EOC outperforms OC when using the same number of options in subsequent tasks. %In particular, the EOC using mixed reward indicated by the brown curve ($\alpha=0.5$) achieves the best performance. 

\begin{comment}
\begin{table}[t]
	\caption{Sample table title}
	\label{sample-table}
	\centering
	\begin{tabular}{lll}
		\toprule
		\multicolumn{2}{c}{Part}                   \\
		\cmidrule{1-2}
		Name     & Description     & Size ($\mu$m) \\
		\midrule
		Dendrite & Input terminal  & $\sim$100     \\
		Axon     & Output terminal & $\sim$10      \\
		Soma     & Cell body       & up to $10^6$  \\
		\bottomrule
	\end{tabular}
\end{table}
\end{comment}
\section{Conclusions}
%\vskip-0.08in
This paper presents a novel algorithm termed eigenoption-critic (EOC) for hierarchical reinforcement learning. EOC extends the option-critic (OC) method by augmenting the extrinsic reward signals with eigenpurposes, intrinsic rewards defined by eigenvectors of the graph Laplacian of the state-space. It allows online eigenoption discovery and it is also applicable to continuous state spaces. Our experiments show that, in general, EOC outperforms OC in both discrete and continuous benchmarks when nonstationarity is introduced. For future work we will test the integration of EOC with deep neural networks on domains such as Atari games~\cite{Bellemare13,Machado17b} and high-dimensional robotic controls~\cite{Duan:2016:BDR:3045390.3045531}.
%\newpage
\bibliography{nips_2017}
\bibliographystyle{abbrvnat}

\end{document}